\documentclass[preprint, twocolumn, 12pt]{IEEEtran}
\usepackage{algorithm}
\usepackage{algorithmicx}
\usepackage{algpseudocode}
\usepackage{stfloats}
\usepackage{ifpdf}
\usepackage{cite}
\usepackage[cmex10]{amsmath}
\usepackage{amssymb}
\usepackage{subfig}
\usepackage{url}
 \usepackage{lineno}
 \usepackage[]{todonotes}
\usepackage{multirow}
\usepackage{leftidx}
\usepackage{color}
\usepackage{soul}
\usepackage{lettrine}
\usepackage{CJK}
\usepackage{array}
\usepackage[colorlinks, citecolor=green]{hyperref}
\usepackage{caption}
\newcommand{\chongyu}[1]{{\color{black} #1}}
\newcommand{\yukai}[1]{{\color{black}#1}}
\newcommand{\keze}[1]{{\color{black}#1}}

\begin{document}

\def\x{{\mathbf x}}
\def\L{{\cal L}}

\author{Yukai~Shi,
    Keze~Wang,
    Chongyu~Chen,
    Li~Xu, and
    Liang~Lin
\thanks{
Yukai Shi, Keze Wang, Chongyu Chen and Liang Lin are with School of Data and Computer Science, Sun Yat-sen University, Guangzhou, China. Keze Wang is also with Dept. of Computing, The Hong Kong Polytechnic
University, Hong Kong. Li Xu is with SenseTime Group Limited, China. Contact: chenchy47@mail.sysu.edu.cn. Liang Lin is also with the Key Laboratory of Machine Intelligence and Advanced Computing, Ministry of Education, Sun Yat-sen University, China.

}}

\title{Structure-Preserving Image Super-resolution via Contextualized Multi-task Learning}

\markboth{IEEE Transactions on Multimedia}%
{Shi \MakeLowercase{\textit{et al.}}:
}

\maketitle
\begin{abstract}
Single image super resolution (SR), which refers to reconstruct a higher-resolution (HR) image from the observed low-resolution (LR) image, has received substantial attention due to its tremendous application potentials. Despite the breakthroughs of recently proposed SR methods using convolutional neural networks (CNNs), their generated results usually lack of preserving structural (high-frequency) details. In this paper, regarding global boundary context and residual context as complimentary information for enhancing structural details in image restoration, we develop a contextualized multi-task learning framework to address the SR problem. Specifically, our method first extracts convolutional features from the input LR image and applies one deconvolutional module to interpolate the LR feature maps in a content-adaptive way. Then, the resulting feature maps are fed into two branched sub-networks. During the neural network training, one sub-network outputs salient image boundaries and the HR image, and the other sub-network outputs the local residual map, i.e., the residual difference between the generated HR image and ground-truth image. On several standard benchmarks (i.e., \textit{Set5}, \textit{Set14} and \textit{BSD200}), our extensive evaluations demonstrate the effectiveness of our SR method on achieving both higher restoration quality and computational efficiency compared with several state-of-the-art SR approaches. \footnote{The source code and some SR results can be found at: \url{http://hcp.sysu.edu.cn/structure-preserving-image-super-resolution/}}
\end{abstract}

\section{Introduction}
\label{sec:intro}
Image super-resolution (SR) is a fundamental problem in image processing.
\yukai{Single image SR approaches, which aim at restoring a high-resolution (HR) image only from a single low-resolution (LR) image, have been applied to many} 
image and video analysis tasks, such as
video surveillance~\cite{SR_app_TMM14},
image-based medical analysis~\cite{SR_App_TMI16}, 
and image/video streaming~\cite{shi2016real,romano2017raisr}.

Common techniques for single image SR can be roughly categorized into reconstruction-, example- and interpolation- based approaches. Reconstruction-based approaches~\cite{irani1991improving,shan2008fast,michaeli2013nonparametric}, which restore HR images by deconvolutional methods~\cite{shan2008fast} with a global blur degradation model, usually introduce ringing artifacts around salient structures~\cite{michaeli2013nonparametric} due to inaccurate blurring kernels in the inverse problem. \yukai{Example-based approaches~\cite{yang2010sc} boost the amplification factor by using internal or external patch data to guide the image restoration. Recently, Huang \textit{et al.}~\cite{huang2015single}}
 proposed to exploit self-similarity for single image SR,
which greatly expands the internal patch searching space. Hu \textit{et al.}~\cite{hu2016serf} proposed a cascaded linear regression technique to model the relationship between HR and LR images. Interpolation-based approaches can achieve acceptable trade-off between performance and efficiency with a pre-defined kernel. However, pre-defined kernels use fixed weights for interpolation, which will inevitably cause blur when the weight definition is inconsistent with image structures. To address issue, various adaptive interpolations~\cite{chu2008gradient,van2012polygon, pisa15PAMI} are proposed. But the improvements in restoration quality are still limited.

The success of deep convolutional neural network (CNN) in computer vision tasks has inspired novel trends in low-level image restoration researches, such as rain/dirt removal~\cite{eigen2013restoring}, noise removal~\cite{jain2009natural}, face hallucination~\cite{wang2014comprehensive,cao2017face}, hashing~\cite{zhang2015bit} and image inpainting~\cite{xie2012image}. \chongyu{Focusing on learning an end-to-end mapping between the LR images and their corresponding HR images, several CNN-based methods~\cite{dong2014srcnn, NIPS2015xuli, wang2015deep,kim2015accurate} have been proposed to perform image SR} in a pure data-driven manner. That is, they directly minimize the mean squared error (MSE) between the predicted and ground-truth images in the training stage.
Although the restoration performance is significantly improved, the structural
inconsistency between the LR input and HR output still exists. 
This is because human visual system is more sensitive to structural changes,
which are difficult to be exploited from MSE-based loss functions.
Recent advances in image SR try to address this
issue~\cite{bruna2015super,ledig2016photo,johnson2016perceptual} by introducing
feature-based perceptive loss functions to the training stage. However, unwanted artifacts and unreal details are also introduced, which make their SR results look unrealistic.

\chongyu{
Considering single image SR is an ill-defined problem, it is necessary to exploit the priors
of natural image to further improve the SR performance.
Motivated by recent advances in deep learning researches
that exploit priors in the form of context information in designing
neural networks~\cite{VideoContex_TMM15, Parsing_Contex_ICCV15},
in this work, we propose to design neutral networks to investigate two types of image
structural information, i.e.,} \textit{global structural information} which corresponds to salient
boundaries in a global perspective and \textit{residual structural information} which contains
noticeable details that are critical to visual quality. 
\chongyu{The success of multi-task learning framework inspires us
to \keze{leverage} such structural information in a unified manner. \keze{For instance, Yang \textit{et al.}~\cite{yang2013feature} proposed to utilize the common knowledge (e.g., feature selection functions) of multiple tasks as supplementary information to facilitate decision making. Considering aforementioned structural information are usually considered as complementary context rather than common knowledge, in this work, we concentrate on complimentary contextualized multi-task learning for structure-preserving single image SR.}
In particular, we propose a deep
joint contextualized multi-task learning framework\keze{, where} three types of image components are imposed as complimentary contexts and jointly learned,
i.e., the base image content, the boundary map, and the residual map.
Besides a convolutional network that learns content-adaptive interpolations to produce
the intermediate base image, we impose an auxiliary task to back-propagate the
global boundary structural context.
Meanwhile, an independent sub-network is introduced to explicitly model
the noticeable details to provide residual structural context.
}

\chongyu{
The major contribution of this work is the proposed contextualized multi-task learning
framework, which is the first attempt to incorporate joint learning
of local, global, and residual contexts into CNNs for single image SR.
Other contributions mainly come from the proposed content-adaptive interpolation
and the sub-networks for capturing complementary image contents, which enables
better trade-off between restoration quality and the number of network parameters. 
}
\yukai{Extensive experiments on several benchmarks datasets (e.g. \textit{Set5}, \textit{Set14}, \textit{BSD500}) demonstrate that the proposed framework shows superior performance to most learning-based approaches in the perspective of both visual quality and quantitative metrics, which
facilitates \keze{the} real-time image SR process.}

\chongyu{
We would like to point out that a preliminary version of this work is reported
in~\cite{This_ICME16}, which coarsely \keze{concatenates} content-adaptive interpolation
and holistic edge context. 
In this paper, we inherit the idea of preserving structures and \keze{refining} the network architecture.
A simple yet powerful sub-network is further employed to capture noticeable
image details for better visual quality.
The whole framework is re-interpreted from the aspect of joint context learning and
multi-task learning.
Besides, more comparisons with state-of-the-art approaches and more detailed \keze{analyses}
of the proposed modules are added to further verify our statements.}

The rest parts of this paper are organized as follows. Section~\ref{sec:Related} briefly reviews existing machine learning-based SR approaches which motivate this work. Section~\ref{sec:proposed} presents the details of the proposed framework, with thorough analysis of every component. Section~\ref{sec:exp} demonstrates the experimental results on several public benchmarks, comparing with state-of-the-art alternatives. Finally, \keze{Section}~\ref{sec:con} concludes this paper.

\section{Related Work}
\label{sec:Related}

\subsection{Interpolation-based image super-resolution}
Interpolation-based approaches typically start from evenly placing the pixels of LR image to the HR grid (the integral coordinates in the HR image domain).
The basic idea of these approaches is to estimate the unknown pixel values in
the HR grid by weighted average of surrounding known pixels.
Considering common pixel changes in a local region can be approximated by continuous functions, 
people have proposed various weight definitions for image interpolation. For example, bilinear 
interpolation is proposed to utilize local linearity, and bicubic interpolation is proposed to exploit the
high-order continuity~\cite{Cubic_TASSP81}. 
However, there are plenty of pixel changes that cannot be described by these pre-defined functions,
especially for regions with rich image structures. In this case, structures will be blurred
due to improper pixel averaging. To address this problem, various adaptive interpolation~\cite{chu2008gradient,van2012polygon} are proposed. For instance, Walt~\textit{et al.}~\cite{van2012polygon} proposed to express polygonal pixel overlap as a linear operator to improve the interpolation performance. But the improvements are still limited.

\subsection{Multi-task learning in image super-resolution}
Decades of researches on multi-task learning have demonstrated that learning multiple correlated tasks simultaneously can significantly improve the performance of the main task~\cite{multitask_ML97, mt2016mm, al2016TMM, mt2016cvpr, yu2017iprivacy}. 
In single image SR, there is also a trend of utilizing multi-task learning.
For example, Yang~\textit{et al.}~\cite{multitask_IWMR11} proposed a multi-task K-SVD 
learning for image SR, in which example image patches are divided into different groups
and K-SVD is applied to every group.
It is shown that simultaneous learning multiple dictionaries can lead to better SR quality.
Liang~\textit{et al.}~\cite{multitaskSR__ICIP15} proposed a multi-task learning framework that jointly
considers image SR process and the image degeneration process.
These works claim that the multi-task learning framework is a feasible way of utilizing priors in learning-based image SR. 

\subsection{Deep learning in image super-resolution}
\chongyu{Recently, deep learning has achieved
significant quality improvements in image SR}.
For example, Dong~\textit{et al.}~\cite{dong2014srcnn} utilized a three-layer fully convolutional
network to learn the non-linear mapping between HR and LR patches,
which has a close relationship to sparse coding. 
\yukai{Ren~\textit{et al.}~\cite{NIPS2015xuli} introduced Shepard CNNs
to facilitate translation variant interpolation, which gives a solution to both inpainting and SR.
Wang~\textit{et al.}~\cite{wang2015deep} proposed a sparse coding based network for
image SR. Based on learned iterative shrinkage and thresholding algorithm(LISTA)~\cite{gregor2010learning}, they employ a set of neural networks to restore images.
Zeng~\textit{et al.}~\cite{zeng2017coupled} proposed a deep autoencoder for SR,}
\chongyu{which explores the consistent representations of HR and LR images and 
demonstrate a superior efficiency compared to similar methods based on sparse
representation. 
Kumar~\textit{et al.}~\cite{TMM_SR_Kumar} studied on several factors that affect the
training phase 
to facilitate learning-based SR with fewer training samples.}
The models of these methods, although being proposed from different aspects, are trained 
to minimize the squared error w.r.t. the ground-truth HR image, which is not necessarily correlated
to good perceptual quality. Bruna~\textit{et al.}~\cite{bruna2015super} referred this problem
as \textit{regression to mean}. Their proposed solution is a conditional generative model,
which demonstrates improvement over visual quality, but with high time cost in both training
and testing.

\yukai{More recently, researchers notice the importance of image details and make various of attempts for exploration. Kim \textit{et al.}~\cite{kim2015accurate,kim2015deeply} further improved the SR quality by different network architectures such as very deep and recursive network structures. However, these methods heavily rely on very deep networks with plenty of parameters. e.g., a 20-layer convolutional neural network~\cite{simonyan2014vgg}. In addition, perceptual losses have
been proposed for CNNs~\cite{bruna2015super,johnson2016perceptual}, which conduct the loss from the image space to high-level feature space of a pre-trained VGG-net~\cite{simonyan2014vgg}. 
At the same time, \keze{Ledig \textit{et al.}}~\cite{ledig2016photo} proposed to apply adversarial network to the task of SR, which results in more image details but lower PSNR score.
More related to our work, there are several attempts to accelerate image SR. By developing a sub-pixel convolutional layer, Shi \textit{et al.}~\cite{shi2016real} used a single model to handle real-time image SR. Similarity, Dong \textit{et al.}~\cite{dong2016sr} applied convolutional layers on LR image and \keze{upscaled} it with deconvolution. They both promise low computational complexity, but there still exists plenty of room for performance improvement.}

\section{Contextualized Multi-task Learning}~\label{sec:proposed}

\begin{figure*}[ht]
\centering
\includegraphics[width=1\textwidth]{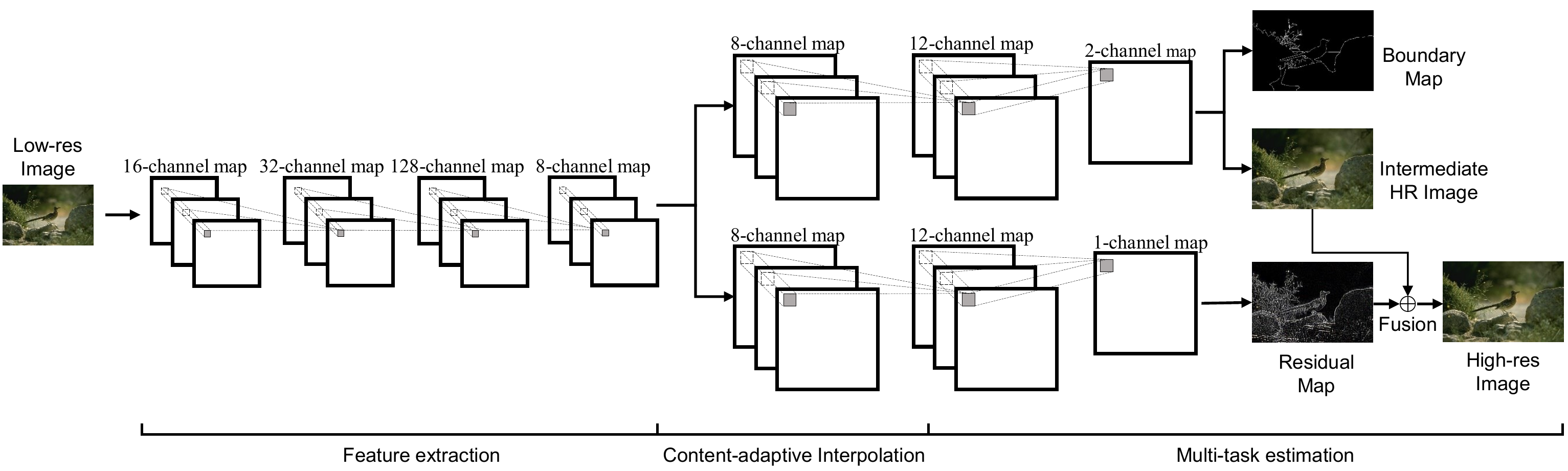}
\caption{ The architecture of our contextualized multi-task deep learning framework for single image super-resolution. Given an input LR image, our framework first extracts its convolutional features and applies one deconvolutional module to interpolate the feature maps in a content-adaptive way. The resulting maps are then fed into two branched sub-networks, which incorporate global boundary context and residual context, respectively. Specifically, during the neural network training, one sub-network outputs salient image boundaries and the intermediate HR image; the other sub-network outputs the local residual map, i.e., the residual difference of the generated HR image and ground-truth image. The final HR estimation is obtained by fusing the intermediate HR image and the local residual map.}
\label{fig:overall-pipeline}
\end{figure*}

In this section, we present \chongyu{the details of our} framework. As sketched in Fig.~\ref{fig:overall-pipeline}, the proposed framework \chongyu{includes} three components: feature extraction, content-adaptive interpolation, and multi-task estimation. 

%
%

\begin{table*}[t] \centering \footnotesize
	\center
	 \vspace{0.05cm}
	 
	\begin{tabular}{c|c|c|c|c|c|c|c|c|c|c}
		\hline
	Component   & \multicolumn{4}{c|}{Feature Extraction} & Interpolation-1 & \multicolumn{2}{c|}{BCN} & Interpolation-2 & \multicolumn{2}{c}{RCN} \\ \hline \hline 
	
    \textbf{\textit{layer}}   & \textit{\textbf{conv}} & \textit{\textbf{conv}} & \textit{\textbf{conv}} & \textit{\textbf{conv}}& \textit{\textbf{deconv}} & \textit{\textbf{conv}} & \textit{\textbf{conv}} & \textit{\textbf{deconv}} & \textit{\textbf{conv}} & \textit{\textbf{conv}} \\ 
    \textbf{\textit{filter}} & 5 & 3 & 3 & 1 & 11 & 3 & 3 & 11 & 3 & 3 \\ 
    \textbf{\textit{channels}} & 16 & 32 & 128 & 8 & 8 & 12 & 2 & 8 & 12 & 1 \\ 
    \textbf{\textit{size}} & 128 & 124 & 124 & 124 & 372 & 372 & 370 & 372 & 372 & 370 \\ 
	\textbf{\textit{parameters}} & 400 & 4,608 & 36,864 & 1,024 & 7,744 & 864 & 216 & 7,744 & 864 & 108 \\ 
	\hline
	\end{tabular}	
	\caption{\chongyu{Detailed setup} \yukai{of each component in our framework. The five rows of the table represent the ``layer type'', ``filter size'', ``output channels'', ``size of output feature maps'' and ``number of parameters'', respectively.}
\chongyu{The content-adaptive interpolation layers for
RCN and BCN are 
``Interpolation-1'' and ``Interpolation-2'', respectively.
Note that this table takes 
the magnification factor of 3 and input images of resolution $128\times128$
as an example of parameter setup. }}
	\vspace{0.1cm}
\label{table:Network_complexity}
\end{table*}

\subsection{Feature Extraction}\label{sec:FE}
\yukai{Inspired by the Pyramid-Net~\cite{han2016deep}, we design a pyramid network
structure for feature extraction.}
\chongyu{That is, there are 3 convolutional layers with 
16, 32 and 128 kernels, respectively.
Detailed setup is summarized in Table~\ref{table:Network_complexity}.
The first layer with kernel size $5\times5$ is designed as a spacious receptive field to capture as much image information as possible, as illustrated in~\cite{he2016deep}.}
\yukai{
The other two layers with $3\times3$ kernel are adopted for better efficiency
as~\cite{chetlur2014cudnn}. Note that we focus on extracting features 
from original LR images instead of the interpolated images.
Thanks to the decreased computations of convolutional operations caused by the small size of feature maps,
the proposed feature extraction can significantly accelerate the speed without
obvious quality drop.
Since the LR image has been represented as high-dimension feature maps through
the first 3 layers, the computation cost may become pretty high if we
import the high-dimension feature maps to content-adaptive interpolation directly.}
\chongyu{
Therefore, we apply a shrinking layer with 8 kernels of size $1\times1$ to reduce
the feature dimension. Note that the kernel number is empirically chosen for a reasonable trade-off between effectiveness and efficiency.}
\yukai{
Benefitting from the shrinking layer, our model not only \keze{avoids} parameter explosion but also \keze{promotes} the restoration efficiency.
}

\begin{figure}[ht]\centering
\subfloat[][ \centering Bicubic kernel \par PSNR: 32.71 dB] {
\includegraphics[width=0.24\textwidth] {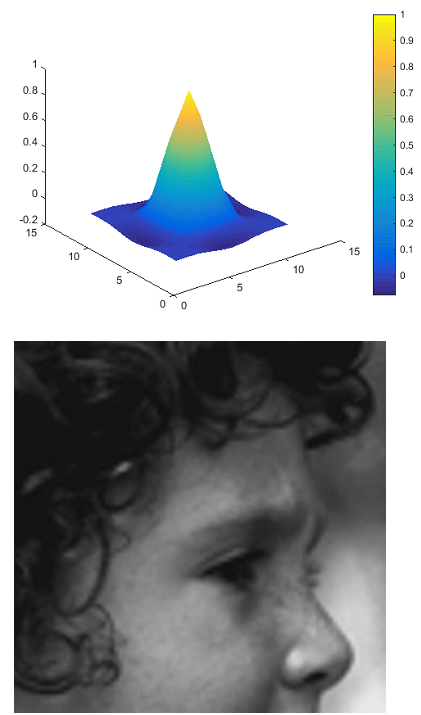}
}
\subfloat[][ \centering Learned kernel \par PSNR: 33.10 dB] {
\includegraphics[width=0.24\textwidth]{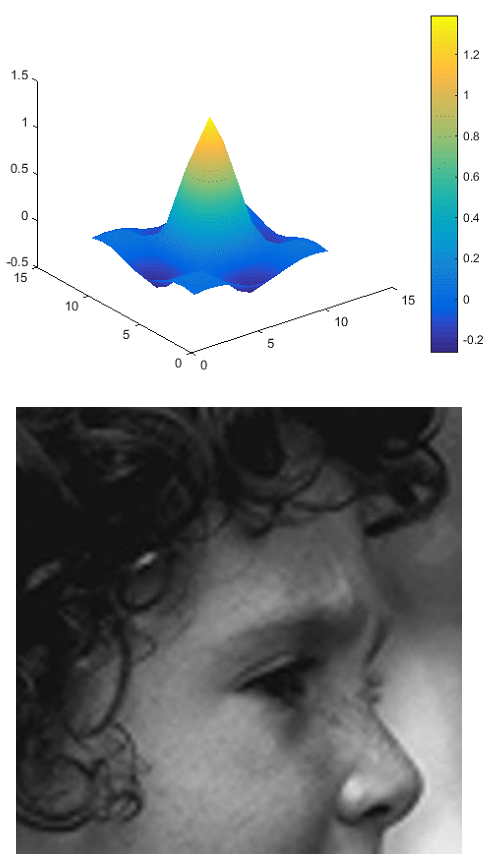}
}
\caption{A comparison between image interpolations by bicubic and learned kernels. }
\label{fig:LSP_Kernels}
\end{figure}

\subsection{Content-adaptive Interpolation}\label{sec:LSPM}
The second component is one deconvolutional layer, which is used to interpolate the LR feature maps in a content-adaptive way. The deconvolutional layer has 8 kernels of size $n \times n$.
Note that in this work, $n$ is determined by the upscaling factor,
which follows the principles of bicubic interpolation.
That is, the kernel should be large enough to cover the second pixel
around the anchor pixel in the HR grid.
For example, the deconvolutional kernel is of size $8\times8$,
$11\times11$, and $16\times16$ for the upscaling factors of
2, 3 and 4, respectively. 
In this way, the deconvolutional layer can be regarded as
a neural network implementation of
standard image interpolation. Let $\mathbf{y}$ be the HR image with a HR grid.
We construct another HR image $\mathbf{x}$ by evenly placed the LR image in 
the HR grid with identical pixel intervals. Then, standard interpolation can be written as:
\begin{equation} \mathbf{y}_j = \sum_{i \in \Omega_j}{ \mathbf{x}_{i}~\omega_{ji}},
\end{equation}
where $i$ and $j$ are the pixel indices in the HR grid, 
$\Omega_j$ represents the subset of $n\times n$ neighbouring pixels around pixel $j$,
and $\omega_{ji}$ is the pre-defined weight for interpolation. 
Note that $ \mathbf{x}_i$ is non-zero only when it comes from a pixel in the LR image.

With these definitions, we re-formulate the interpolation process
as a basic component of a deconvolutional layer, i.e.,
\begin{equation}
\mathbf{y}_j = \delta (\sum_{i\in \Omega_j} \mathbf{x}_i W(i') + b)
\end{equation}
where $\delta(\cdot)$ represents the activation function, $W$ is the deconvolutional kernel,
$i'$ represents the pixel of $W$ that contributes to pixel $j$, and $b$ is the bias.

In the proposed content-adaptive interpolation, we use multiple deconvolutional kernels in a similar fashion. 
That is, we evenly place the LR image in the HR grid to construct $\mathbf{h}^l$. Then,
\begin{equation} 
\mathbf{h}^{l+1}_k = \delta ( \mathbf{h}^{l} \otimes W_k
+ b_k ), \end{equation}
where the subscript $k$ represents the kernel index, ``$\otimes$'' represents the
convolutional operator, and $\mathbf{h}^{l+1}$ is the output image of the
$l^{th}$ layer.
In this way, content-adaptive image interpolation can be accomplished via a deconvolutional layer,
whose kernels are learned from sufficient training data.
\yukai{Note that the deconvolutional layer is in the middle of the proposed network, which is different from other CNN-based SR methods~\cite{NIPS2015xuli, dong2014srcnn} that use deconvolution as the last layer. It is shown empirically that the proposed network can achieve nice restoration quality with reasonably increasing network parameters.}

\yukai{To compare the proposed network with the bicubic interpolation, we construct a small network which only has one deconvolutional layer to learn an adaptive kernel, taking BSD300 as training data and bicubic interpolation parameters for initialization. The intensity changes of bicubic and our learned kernels are visualized in Fig.~\ref{fig:LSP_Kernels}, which illustrates that the learned kernel contains more high-frequency components. Meanwhile, the restoration results also indicate that the learned kernel leads to a superior restoration quality with more recovered details compared to the bicubic kernel. Thus, the effectiveness of the proposed adaptive interpolation is verified.}

\subsection{Contextualized Multi-task Learning}\label{sec:Multi}

\begin{figure} [t]\centering
\subfloat[][ \centering original images] {
\includegraphics[width=0.22\textwidth] {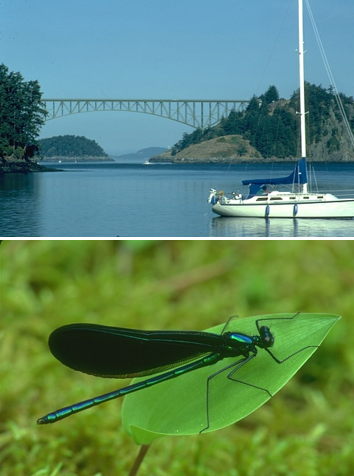}
}
\subfloat[][ \centering boundary maps] {
\includegraphics[width=0.22\textwidth]{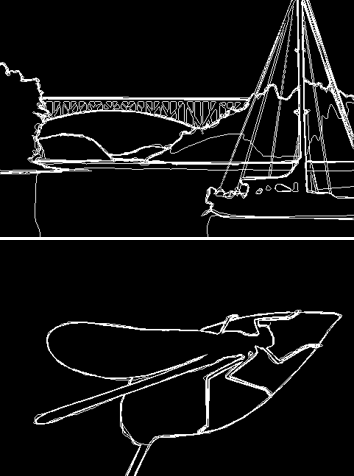}
}
\caption{Example images with salient boundaries. (a) Original images. (b) Manually labeled edge
maps.}
\label{fig:HSP-boundaries}
\end{figure}
In spired by the multi-task learning principles, we make an attempt to introduce auxiliary knowledge to SR issue.

\begin{figure}[t] \centering
\centering
\includegraphics[width=0.115 \textwidth]{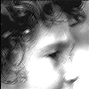}
\includegraphics[width=0.115 \textwidth]{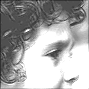}
\includegraphics[width=0.115 \textwidth]{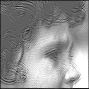}
\includegraphics[width=0.115 \textwidth]{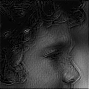}
\includegraphics[width=0.115 \textwidth]{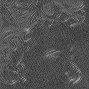}
\includegraphics[width=0.115 \textwidth]{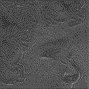}
\includegraphics[width=0.115 \textwidth]{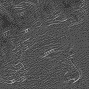}
\includegraphics[width=0.115 \textwidth]{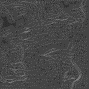}
\caption{Illumination of several representative feature maps produced by the first three
layers of feature extraction. The top row and bottom row show image-like and edge-like features, respectively.
}
\label{fig:LR_fmap}
\end{figure}

\textit{\textbf{Global Boundary Context}}: We develop a Boundary Context sub-Network (BCN) to preserve salient boundaries that represent global image structures.
BCN consists of two convolutional layers with $3\times3$ kernels, where one layer is with 12 kernels and the other layer is with 2 kernels. In the training phase of BCN, we propose to exploit salient image boundaries by regarding edge detection as a joint task of HR image restoration. In particular, we introduce an auxiliary term into the objective function, which computes the error between predicted and human-labeled edge/boundary maps. These boundary maps are from Berkeley Segmentation Dataset (BSD)~\cite{amfm_pami2011}. Note that there are multiple boundary maps in BSD500 data set, we use their summation for better visualization and \keze{show} the examples in Fig.\ref{fig:HSP-boundaries}. 

With the two tasks of image restoration and edge detection, image components and
structural features are firstly extracted and enlarged by content-adaptive interpolation before being fed into the BCN. Several representative samples of the extracted feature maps are shown in Fig.~\ref{fig:LR_fmap}, in which the top row and bottom row show image-like and edge-like features, respectively. This implies that these layers simultaneously extract redundant components and features, making it possible to produce base image and boundary maps in the HR image domain. 

Through joint optimization in an end-to-end manner, feature extraction, content-adaptive interpolation and BCN can provide complimentary context information to each other. In this way, structure-aware feature representations can be learned with the content-adaptive interpolation. 

\textit{\textbf{Residual Context}}: As a result of paying close attention to generating the HR image with salient boundaries, the concatenated BCN might fail to restore some subtle but noticeable structures. Motivated by the recent residual learning paradigm~\cite{kim2015accurate, csc_sr}, we make an attempt to address this issue by employing a Residue Context sub-Network (RCN). The objective of the RCN is to synthesize a residual image, which is defined as the difference between the interpolated HR image and the ground-truth HR image. In contrast to using the bicubic interpolated HR image as in \cite{kim2015accurate} and \cite{csc_sr}, our model uses the intermediate HR image provided by BCN. This can bring us two benefits: i) Higher image SR performance. As the HR image provided by BCN achieves comparable performance to the state-of-the-art methods, RCN can focus on remedying the overlooked information for higher SR quality; ii) A lightweight network architecture for RCN. Our used interpolated image contains significantly richer information than the bicubic one. Hence, compared with \cite{kim2015accurate} and \cite{csc_sr}, the synthesization of residual images is much easier. As illustrated in Fig.~\ref{fig:overall-pipeline}, the architecture of RCN is the same as that of the concatenated BCN.

For the joint optimization of content-adaptive interpolation, BCN and RCN, we develop a fusion layer to merge the intermediate output of RCN and BCN in a data-driven way. In particular, the final HR image $\mathbf{y}$ of our framework is obtained by:
\begin{equation}
\mathbf{y}=\mathbf{f} \otimes \mathbf{I}_{interHR} + \mathbf{I}_r,
\label{equation:the-whole-obj2}
\end{equation}
where $\mathbf{f}$ denotes a $3\times3$ convolutional filter, $\mathbf{I}_{interHR}$ is the intermediate HR image provided by BCN, and $\mathbf{I}_r$ is the residue image synthesized by RCN. In this way, the parameters of $\mathbf{f}$ can be adaptively \keze{updated} during the learning process.

\section{Framework Training}
The proposed framework is jointly optimized on a set of ``LR image, HR image and HR edge
map\footnote{In BSD data sets, more than one boundary maps are provided for every image, which
are all used in our training process. Since multiple boundary maps are used in the same way, in this
subsection, we focus on the case of one boundary map for simplicity.}'' triplets.
For convenience, we use $\mathbf{I}_l$,  $\mathbf{I}_h$ and $\mathbf{I}_b$ to represent
the LR image, HR image and boundary map, respectively.
Given the input $\mathbf{I}_l$, the objective of our model is to reconstruct a HR image similar to $\mathbf{I}_h$ and predict a boundary map similar to $\mathbf{I}_b$.

The parameter $\mathbf{W}$ of our model can be divided into 4 disjoint parts, i.e., $\mathbf{W}=\{\mathbf{W}_{s}, \mathbf{W}_{h}, \mathbf{W}_{b}, \mathbf{W}_{d}\}$, where $\mathbf{W}_{s}$ and $\mathbf{W}_{d}$ denote the parameters of content-adaptive interpolation and RCN, respectively. We denote the parameter of feature extraction stage has combined into content-adaptive interpolation part. For BCN, we use $\mathbf{W}_{h}$ and $\mathbf{W}_{b}$ to represent the specific weights for generating the intermediate HR image and the boundary maps, respectively.
Since the parameters are separable, we propose to train our model in three iterative steps.
First, we jointly train content-adaptive interpolation and BCN until their convergence; Second, fixing the parameters of content-adaptive interpolation and BCN, we update the parameters of RCN. Third, we jointly optimize content-adaptive interpolation, BCN and RCN. Specifically, content-adaptive interpolation and BCN are trained according to the following objective function:
\begin{equation}
\begin{array}{rl}
L(\mathbf{I}_l,\mathbf{I}_h, \mathbf{I}_b, \mathbf W)  = & L_{h}(\mathbf{I}_{l},\mathbf{I}_{h}, \mathbf W_{s}, \mathbf W_{h})  +\\ & \alpha \cdot L_{b} (\mathbf{I}_l,\mathbf{I}_b, \mathbf W_{s}, \mathbf W_{b}),
\end{array}
\label{equation:the-whole-obj}
\end{equation}
where $L_{h}$ and $L_{b}$ represent the HR image reconstruction objective and the boundary prediction objective, respectively.
The balance weight $\alpha$ is used to control the importance of $L_{h}$ and $L_{b}$, which is empirically set to 1 in all our experiments.
Both $L_{h}$ and the $L_{b}$ are in the form of mean squared error (MSE), i.e.,
\begin{equation}
L_{h} = \frac{1}{N}\sum_{i=1}^{N}\left ( \mathbf{I}_h^{i} - f_{h}( \mathbf{W}_{s}, \mathbf{W}_{h}, \mathbf{I}_l^i) \right ) ^ 2,
\label{equation:the-hr-obj}
\end{equation}
and 
\begin{equation}
L_{b} = \frac{1}{N}\sum_{i=1}^{N}\left ( \mathbf{I}_b^{i} - f_{b}( \mathbf{W}_{s}, \mathbf{W}_{b}, \mathbf{I}_l^i) \right ) ^ 2, 
\label{equation:the-boundaries-obj}
\end{equation}
where $f_{h}(\cdot)$ and 
$f_{b}(\cdot)$ 
denote the reconstructed HR image and the predicted boundary map, respectively,
$i$ represents the sample index, and $N$ is the number of training triplets.
For simplicity, we use $\mathbf{I}_\omega$ to denote $f_{b}( \mathbf{W}_{s}, \mathbf{W}_{b}, \mathbf{I}_l)$.
Note that when multiple boundary maps are available, there will be more edge prediction objectives.

\begin{algorithm}[t]
\caption{Contextualized Multi-task Learning.} 
\label{alg}
\label{alg:batchTraining}
\begin{algorithmic}[1]
\raggedright

\Require 
Training LR images $I_l$; HR images $I_h$; boundary images $I_b$;
\While    {$t<T$}
\State $t\leftarrow t+1$;
\State Randomly select a subset of LR images, HR images and boundary images $\mathbf{I}'_l,\mathbf{I}'_h,\mathbf{I}'_b$ from the training set;
\State \textbf{for all} {$\mathbf{I}_l^{'i}$} \textbf{do}
\State Obtain $f_{h}(\mathbf{W}_{s}, \mathbf{W}_{h}, \mathbf{I}_l^{'i})$ and $f_{b}(\mathbf{W}_{s}, \mathbf{W}_{b}, \mathbf{I}_l^{'i})$ via forward propagation;
\State Update $\mathbf{W}_s^t, \mathbf{W}_h^t, \mathbf{W}_b^t$ via the intermediate HR output and boundary output: $\frac {\partial L_h}{\partial f_{h}(\mathbf{W}_{s}, \mathbf{W}_{h}, \mathbf{I}_l^{'i})}$,$\frac {\partial L_b}{\partial f_{b}(\mathbf{W}_{s}, \mathbf{W}_{b}, \mathbf{I}_l^{'i})}$; 
\State \textbf{end for}
\EndWhile
\While    {$t<2T$}
\State $t\leftarrow t+1$;
\State \textbf{for all} {$\mathbf{I}_l^{'i}$} \textbf{do}
\State Obtain $f_{d}(\mathbf{W}_{s}, \mathbf{W}_{d}, \mathbf{I}_l^{'i})$ via forward propagation;
\State Update $\mathbf{W}_d^t$ via the residual output and intermediate HR output: $\frac {\partial L_d}{\partial( f_{d}(\mathbf{W}_{s}, \mathbf{W}_{d}, \mathbf{I}_l^{'i}) + f_{h}(\mathbf{W}_{s}, \mathbf{W}_{h}, \mathbf{I}_l^{'i}))}$; 
\State \textbf{end for}
\EndWhile
\end{algorithmic} 
\end{algorithm}

The loss function for training RCN is defined as:
\begin{equation}
L_d =\frac{1}{N}\sum_{i=1}^{N}(\mathbf{I}_h^i-\mathbf{I}_\omega^i-f_d(\mathbf{W}_s,\mathbf{W}_d, \mathbf{I}_l^i))^2.
\label{equation:the-whole-obj3}
\end{equation}

Finally, the whole framework is optimized by employing the standard back propagation algorithm, i.e., 
\begin{equation}
L = \frac{1}{N} \sum_{i=1}^N (\mathbf{I}_h^i - y)^2,
\end{equation}
where $y$, the output of fusion layer, is the final HR image in the testing phase.

The whole training phase is summarized as Algorithm~\ref{alg},
which accords with the pipeline of our proposed framework in Fig.~\ref{fig:overall-pipeline}.

\section{Experiments}\label{sec:exp}

\subsection{Experiment Setting}
\indent \textit{\textbf{Datasets}}: All experiments are evaluated on three challenging benchmarks, i.e., \textit{Set5}~\cite{bevilacqua2012low},
\textit{Set14}~\cite{zeyde2012single} and \textit{BSD500}~\cite{amfm_pami2011}. The \textit{BSD500} dataset consists of 500 natural images and human annotations for corresponding boundaries. We use the 300 images \keze{from} its training and validation set for training.
The rest of 200 images in \textit{BSD500} dataset form a widely used benchmark called \textit{BSD200}. Besides, the \textit{Set5} and \textit{Set14} datasets are also adopted
as testing sets in other state-of-the-art methods such as~\cite{dong2014srcnn,wang2015deep,kim2015accurate}. Thus, we conduct experiments on the three
benchmarks.

\begin{table*}[t] \centering 
\begin{center}
\begin{tabular}{|c|*{10}{>{\hfil}p{33pt}<{\hfil}| }}
\hline
Test set & \multicolumn{3}{c|}{Set5} & \multicolumn{3}{c|}{Set14} & \multicolumn{3}{c|}{BSD200} \\
\hline
Scaling factor & $\times2$ & $\times3$ & $\times4$ & $\times2$ & $\times3$ & $\times4$ & $\times2$ & $\times3$ & $\times4$\\
\hline \hline
Bicubic & 33.66 & 30.39 & 28.42 & 30.23 & 27.54 & 26.00 & 29.43 & 27.18 & 25.92\\
A+~\cite{a+} & 36.55 & 32.59 & 30.28 & 32.28 & 29.13 & 27.32 & 31.44 & 28.36 & 26.83\\
SRCNN~\cite{dong2014srcnn} & 36.34 & 32.59 & 30.09 & 32.18 & 29.00 & 27.20 & 31.38 & 28.28 & 26.73 \\
SRF~\cite{schulter2015fast} & 36.89 & 32.72 & 30.35 & 32.52 & 29.23 & 27.41 & 31.66 & 28.45 & 26.89\\
FSRCNN~\cite{dong2016sr} & \underline{36.94} & 33.06 & 30.55 & 32.54 & 29.37 & 27.50 & 31.73 & 28.55 & 26.92 \\
SCN~\cite{wang2015deep} & 36.93 & \underline{33.10} & \underline{30.86} & \underline{32.56} & \underline{29.41} & \underline{27.64} & 31.63 & 28.54 & \underline{27.02} \\   
ShCNN~\cite{NIPS2015xuli} & 36.83 & 32.88 & 30.46 & 32.48 & 29.39 & 27.51 & \underline{31.75} & \underline{28.60} & 26.95 \\ \hline
Proposed & \textbf{37.17} & \textbf{33.45} & \textbf{31.11}& \textbf{32.77} & \textbf{29.63} & \textbf{27.79} & \textbf{31.81} & \textbf{28.67} & \textbf{27.11} \\
\hline
\end{tabular}
\end{center}
\caption{Quantitative comparisons among different methods in terms of PSNR (dB),
in which the underline indicates the second place and bold face represents the first place.}
\label{table:PSNR-on-different-method}
\end{table*}

\begin{table*}[t] \centering 
\begin{center}
\begin{tabular}{|c|*{10}{>{\hfil}p{33pt}<{\hfil}| }}
\hline
Test set & \multicolumn{3}{c|}{Set5} & \multicolumn{3}{c|}{Set14} & \multicolumn{3}{c|}{BSD200} \\
\hline \hline
Scaling factor & $\times2$ & $\times3$ & $\times4$ & $\times2$ & $\times3$ & $\times4$ & $\times2$ & $\times3$ & $\times4$\\
\hline
Bicubic & 0.9299 & 0.8682 & 0.8104 & 0.8687 & 0.7736 & 0.7019 & 0.8524 & 0.7469 & 0.6727\\
A+~\cite{a+} & 0.9544 & 0.9088 & 0.8603 & 0.9056 & 0.8188 & 0.7491 & 0.8966 & 0.7945 & 0.7171\\
SRCNN~\cite{dong2014srcnn} & 0.9521 & 0.9033 & 0.8530 & 0.9039
 & 0.8145 & 0.7413 & 0.8835 & 0.7794 & 0.7018 \\
SRF~\cite{schulter2015fast} & 0.9536 & 0.9046 & 0.8529 & 0.9042 & 0.8168& 0.7457 & 0.9011 & 0.8053 & 0.7332\\
FSRCNN~\cite{dong2016sr} & 0.9552 & \underline{0.9128} & 0.8619 & 0.9080 & 0.8231 & 0.7509 & 0.9064 & 0.8123 & 0.7378 \\
SCN~\cite{wang2015deep} & \underline{0.9571} & 0.9112 & \underline{0.8644} & \underline{0.9093} & \underline{0.8246} & \underline{0.7541} & 0.9058 & 0.8139 & 0.7403 \\
ShCNN~\cite{NIPS2015xuli} & 0.9551 & 0.9109 & 0.8638 & 0.9079 & 0.8239 & 0.7530 & \underline{0.9069} & \underline{0.8144} & \underline{0.7407} \\
 \hline 
Proposed & \textbf{0.9583} & \textbf{0.9175} & \textbf{0.8736}& \textbf{0.9109} & \textbf{0.8269} & \textbf{0.7594} & \textbf{0.9074} & \textbf{0.8182} & \textbf{0.7460} \\
\hline
\end{tabular}
\end{center}
\caption{Quantitative comparisons among different methods in terms of SSIM,
in which the underline indicates the second place and bold face represents the first place.}
\label{table:SSIM-on-different-method}
\end{table*}

\textit{\textbf{Implementation details}}: In the training phase, we first convert the original color image to grayscale image by extracting the luminance component in YCbCr color space.
Then, we downscale the training images by requested scaling factors (e.g., 2, 3, and 4) to obtain the LR images. The LR images are cropped into a set of patches with a stride of 4. { The size of patches \keze{is} set to be same as receptive field.}
The corresponding HR images and boundary maps are cropped with respect to the scaling factors.
Before training, we initialize the network parameters by a zero-mean Gaussian distribution with a standard deviation of $1\times10^{-4}$. 
For the pre-training of the proposed model, we use the 91-images~\cite{yang2010sc}
and PASCAL VOC2012~\cite{voc2012} datasets, which totally contain 13,487 clear images.
Specifically, the model using LR and HR image pairs is pre-trained following the same strategy as~\cite{dong2014srcnn}. Since the feature extraction stage employ pyramid structure, we speed it up with the help of Factorized CNN~\cite{Wang2016Factorized}. In the training on \textit{BSD300} dataset, The learning rate of the last layer is set to $1\times10^{-5}$, while the rest layers are using a fixed learning rate of $1\times10^{-4}$.
To increase the number of training samples, we also employ data augmentation for \textit{BSD300}
dataset, as reported in~\cite{wang2015deep}.

\begin{table}[tbp] \centering \small
\center
\begin{tabular}{|*{4}{c|}}
\hline
Methods & Parameter number & PSNR \\
\hline\hline
SRCNN \cite{dong2014srcnn} & 57,184 & 32.59 \\
FSRCNN \cite{dong2016sr}& 15,740 & 33.06 \\
VDSR \cite{kim2015deeply}& 664,704 & 33.66 \\
\hline 
Ours & 60,436 & 33.45 \\
Deeper ours & 594,964 & \textbf{33.80}\\
\hline
\end{tabular}
\caption{Comparison on parameter number and PSNR performance on \textit{Set5} with a scaling factor of 3.}
\label{table:Parameter_Comparison}
\end{table}

\begin{figure} [tbp]
\centering
\includegraphics[width=1 \columnwidth]{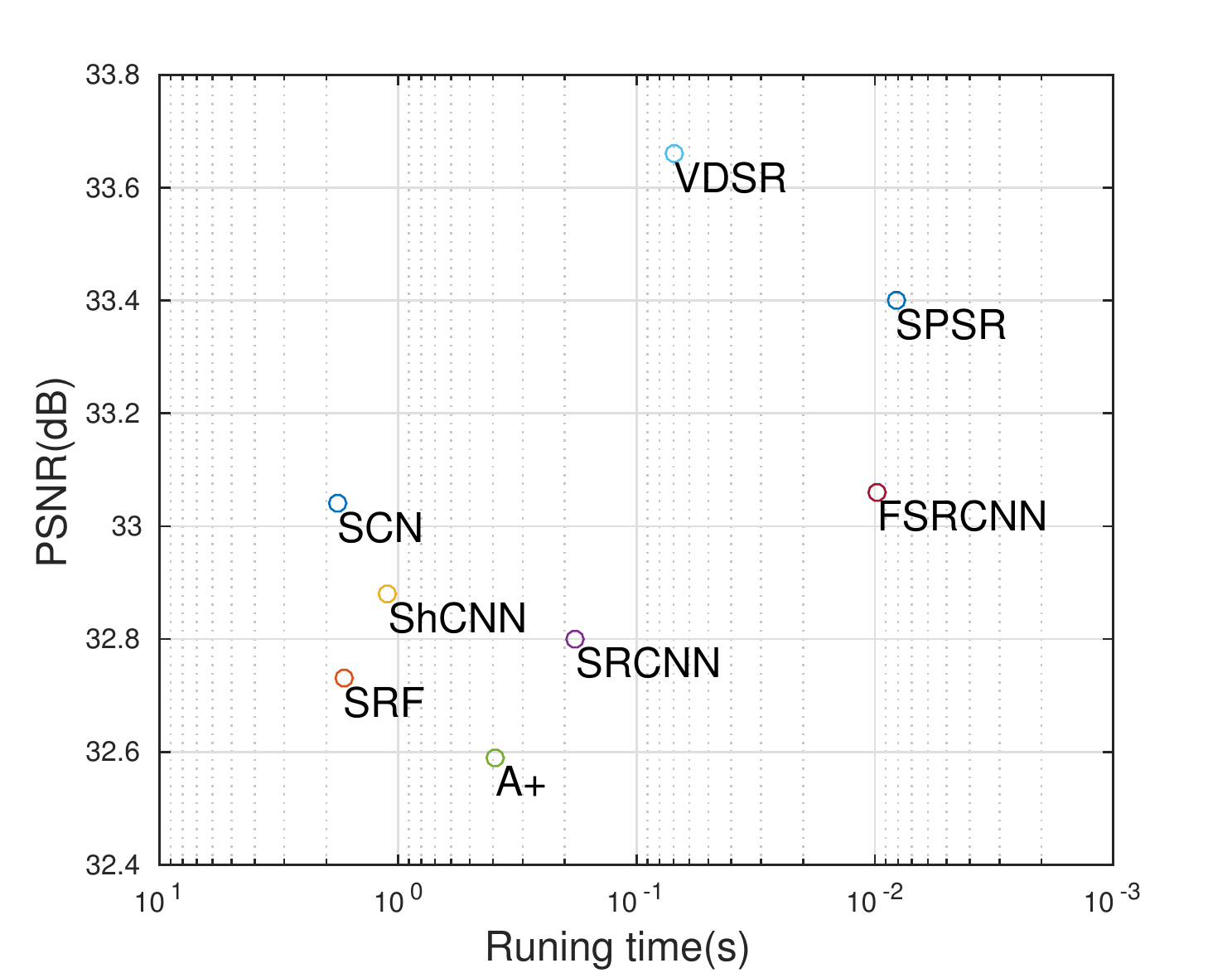}
\caption{The efficiency analysis for the scaling factor of 3 on the \textit{Set5} dataset.
}
\label{fig:efficiency_comparison}
\end{figure}

\textit{\textbf{Methods and metrics}}: We compare our model with several recent state-of-the-art methods, including a three-layer CNN (SRCNN)~\cite{dong2014srcnn}, super-resolution forest (SRF)~\cite{schulter2015fast}, sparse coding-based network (SCN)~\cite{wang2015deep}, anchored neighborhood regression (A+)~\cite{kim2015accurate}, shepard interpolation neural network (ShCNN)~\cite{NIPS2015xuli}, very deep convolutional network (VDSR)~\cite{kim2015accurate}, and fast convolutional network for SR (FSRCNN)~\cite{dong2016sr}.
For fair comparisons, we employ the popular PSNR and SSIM metrics for evaluation.
To evaluate the structure-preserving capability, we introduce a new metric called ``EPSNR'', which can be formulated as:
\begin{equation}
EPSNR = 10\log_{10} {\left(\frac{MAX_I^2}{\frac{1}{|E|}\sum\limits_{i \in E}{\left(G_i-P_i\right)^2}}\right)},
\end{equation}
where $MAX_I=255$ is used for 8-bit images, $G$ and $P$ denote the ground-truth and the produced HR images, respectively, $E$ indicates the pixels whose distances to their closest boundary are less than 2 pixels, and $i$ is the pixel index. It is believed that EPSNR can better exploits image fidelity on edge regions.

\begin{figure*}[htbp]\centering
\subfloat[][ \centering Bicubic \par 26.64 / 0.8232] {
\includegraphics[width=0.21\textwidth]{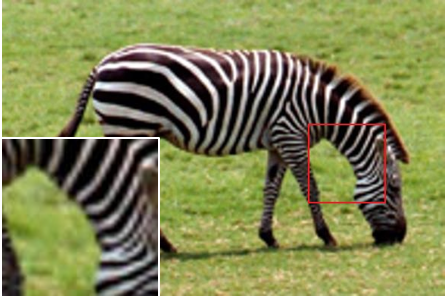}
}
\subfloat[][ \centering A+ \cite{a+} \par 29.11 / 0.8462] {
\includegraphics[width=0.21\textwidth]{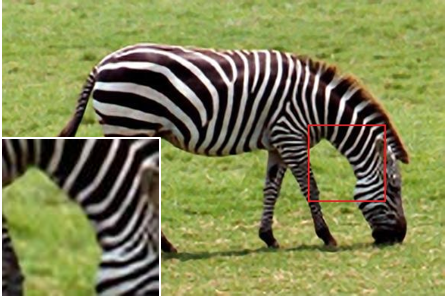}
}
\subfloat[][ \centering SRF \cite{schulter2015fast} \par 29.23 / 0.8483] {
\includegraphics[width=0.21\textwidth]{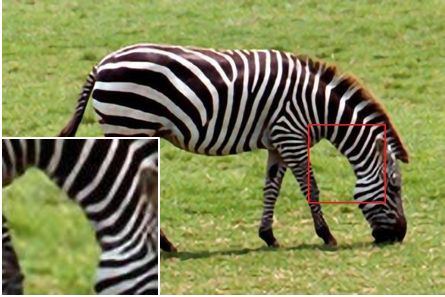}
}
\subfloat[][ \centering SRCNN ~\cite{dong2014srcnn}\par 29.34 / 0.8513] {
\includegraphics[width=0.21\textwidth]{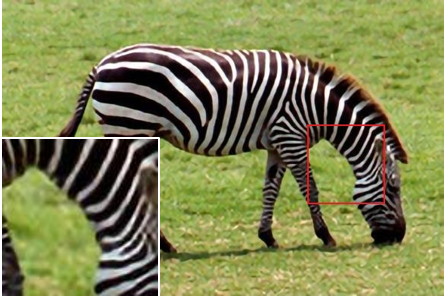}
}
\\
\subfloat[][ \centering SCN ~\cite{wang2015deep}\par 29.58 / 0.8499 ] {
\includegraphics[width=0.21\textwidth]{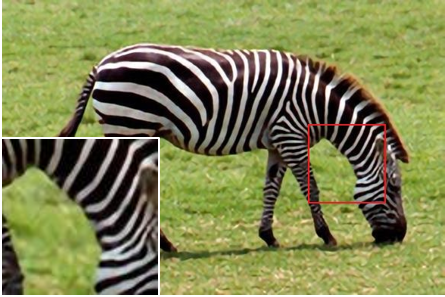}
}
\subfloat[][ \centering ShCNN ~\cite{NIPS2015xuli}\par 29.61 /  0.8521 ] {
\includegraphics[width=0.21\textwidth]{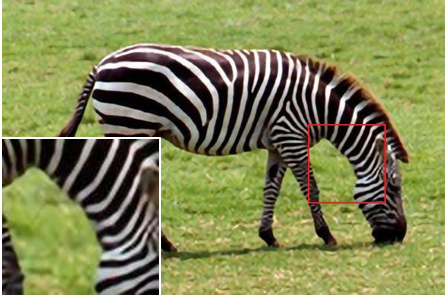}
}
\subfloat[][ \centering Proposed \par \textbf{29.80} / \textbf{0.8589} ] {
\includegraphics[width=0.21\textwidth]{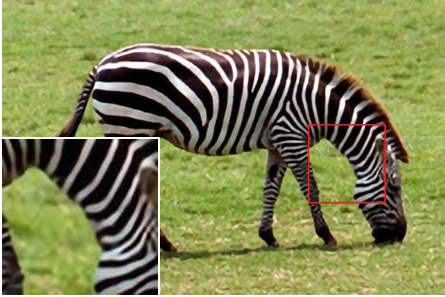}
}
\subfloat[][ \centering Original \par PSNR / SSIM ] {
\includegraphics[width=0.21\textwidth]{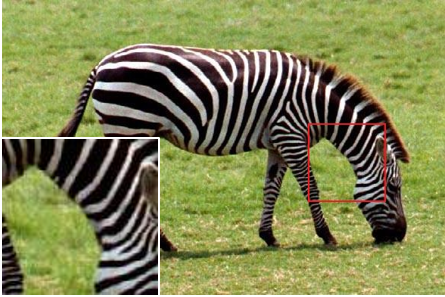}
}
\caption{\yukai{Visual comparison on the ``Zebra'' image from \textit{Set14} (factor 3), where the PSNR
and SSIM are separated by ``/''.}}
\label{fig:set5-visualize}

\subfloat[][ \centering Bicubic \par 22.18 / 0.7376] {
\includegraphics[width=0.21\textwidth] {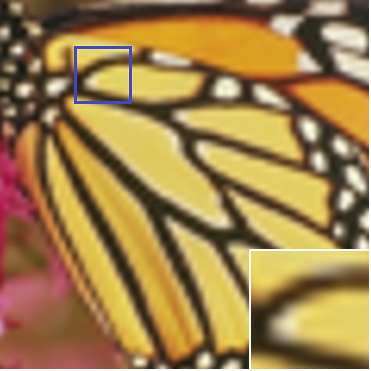}
}
\subfloat[][ \centering A+ \cite{a+} \par 24.68 / 0.8402] {
\includegraphics[width=0.21\textwidth] {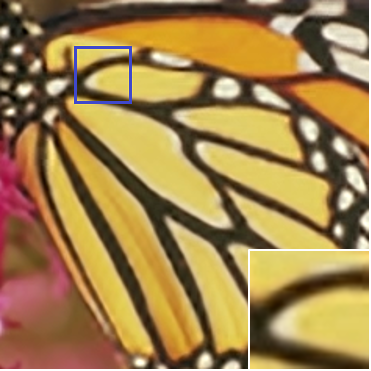}
}
\subfloat[][ \centering SRF \cite{schulter2015fast} \par 24.60 / 0.8280] {
\includegraphics[width=0.21\textwidth] {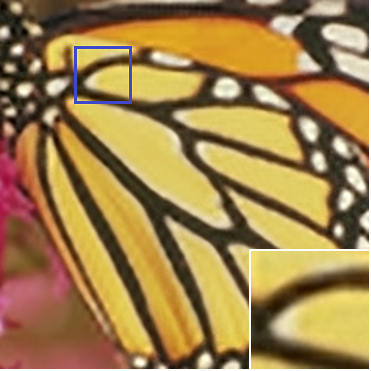}
}
\subfloat[][ \centering SRCNN \cite{dong2014srcnn} \par 25.31 / 0.8677 ] {
\includegraphics[width=0.21\textwidth] {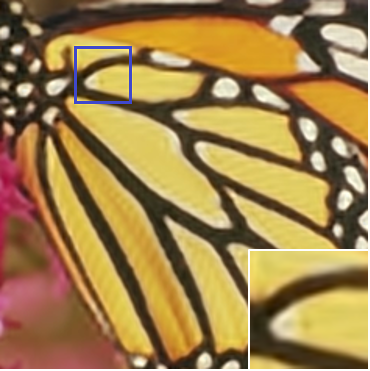}
}
\\
\subfloat[][ \centering SCN \cite{wang2015deep} \par 25.98 / 0.8821 ] {
\includegraphics[width=0.21\textwidth] {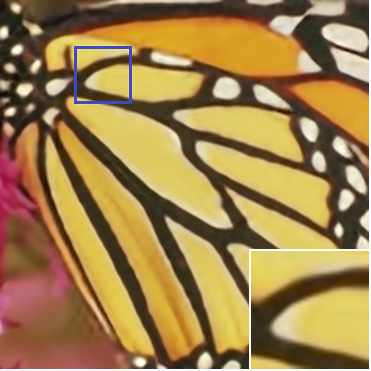}
}
\subfloat[][ \centering ShCNN \cite{NIPS2015xuli} \par 25.85 /  0.8677  ] {
\includegraphics[width=0.21\textwidth] {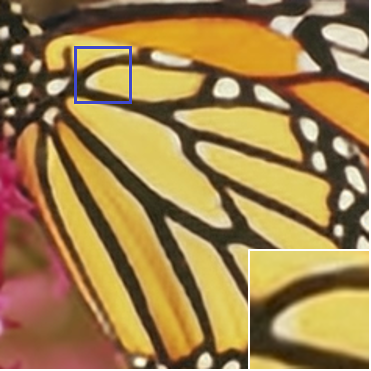}
}
\subfloat[][ \centering Proposed \par \textbf{26.05} / \textbf{0.8830} ] {
\includegraphics[width=0.21\textwidth] {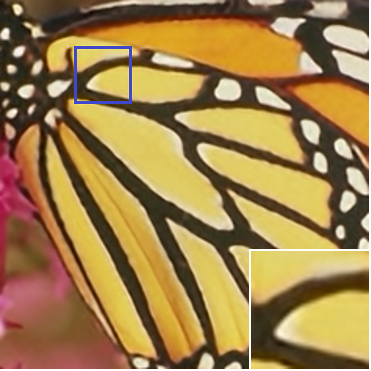}
}
\subfloat[][ \centering Original \par PSNR / SSIM ] {
\includegraphics[width=0.21\textwidth] {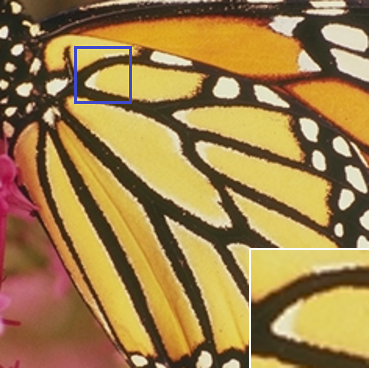}
}
\caption{Visual comparisons on the ``Butterfly'' image from \textit{Set5} (factor 4), where the PSNR
and SSIM are separated by ``/''.}
\label{fig:internet-visualize}
\end{figure*}

\begin{figure}[t] \centering
\centering

\subfloat[][ \centering Bicubic \par 21.51 dB] {
\includegraphics[width=0.14\textwidth] {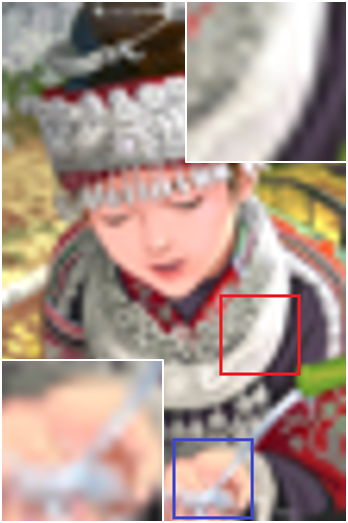}
}
\subfloat[][ \centering ShCNN\cite{NIPS2015xuli} \par 22.54 dB ] {
\includegraphics[width=0.14\textwidth]{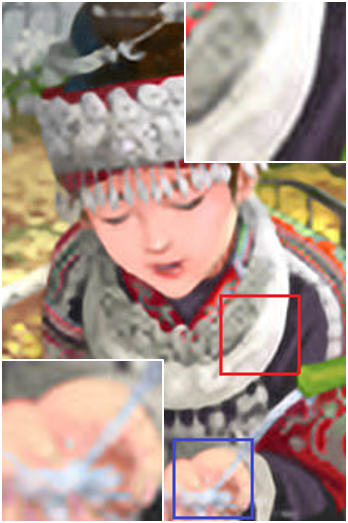}
}
\subfloat[][ \centering SRGAN-1~\cite{ledig2016photo} \par 20.45 dB] {
\includegraphics[width=0.14\textwidth]{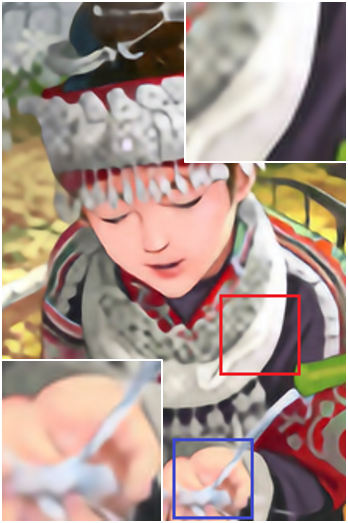}
}
\\
\subfloat[][ \centering SRGAN-2~\cite{ledig2016photo} \par 19.07 dB] {
\includegraphics[width=0.14\textwidth]{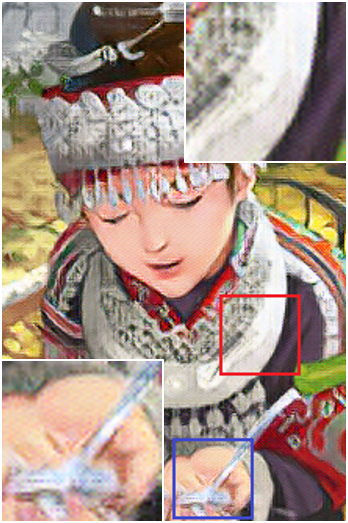}
}
\subfloat[][ \centering Proposed \par 22.72 dB] {
\includegraphics[width=0.14\textwidth]{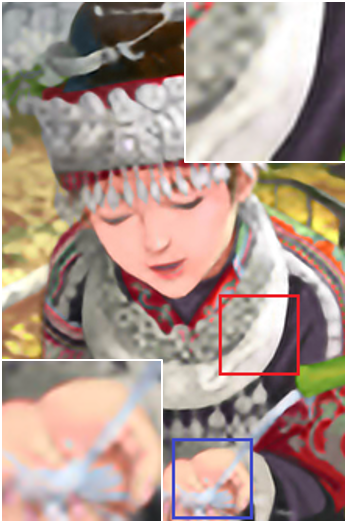}
}
\subfloat[][ \centering Ground Truth ] {
\includegraphics[width=0.14\textwidth]{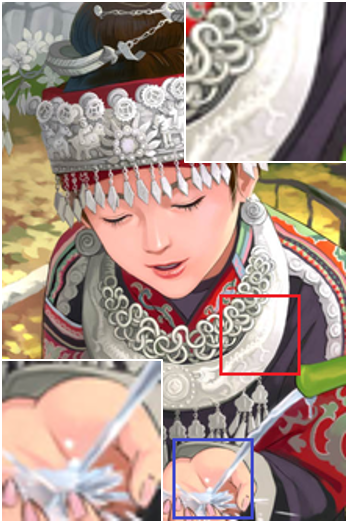}
}
\caption{Visual comparison on Bicubic, ShCNN, our proposed and SRGAN methods. Note that, \keze{`SRGAN-1'} represents the adversarial network with MSE-based content loss only. \keze{`SRGAN-2'} is the adversarial network with perceptual loss as mentioned in~\cite{ledig2016photo}.
}
\label{fig:GAN}
\vspace{-8pt}
\end{figure}

We have also investigated the model complexity from the aspect of parameter number. Two profiles of our model are used, i.e., the common model (denoted as ``ours'') used in the above comparisons, and the model with a much deeper architecture (denoted as ``deeper ours''). In the ``deeper ours'' profile, we only increase the convolutional layer number of feature extraction stage from 4 to 18. Thus our model has similar number of parameters compared to VDSR. Both profiles can be accelerated by cuDNN~\cite{chetlur2014cudnn}. All the CNN-based methods are compared using the \textit{Set5} dataset with a scaling factor of 3. The results illustrated in Table~\ref{table:Parameter_Comparison} \keze{demonstrate} that the performance of our model keeps increasing as the parameter number increases. Using comparable network parameters, our model can achieve a PSNR gain of 0.14~dB compared to VDSR. 
Since fewer parameters can benefit both the training and testing phases, we recommend our model with the common profile. Fig.~\ref{fig:efficiency_comparison} illustrates the efficiency of all the compared methods using the ``time-quality'' diagram. It is demonstrated that our model with common profile runs nearly 2 times faster than VDSR while maintaining the second best SR performance, which is quite suitable for lightweight and fast implementation on consumer-grade devices.
For applications that require extremely high SR quality, deeper ours will be a nice choice.

\begin{figure}[t] \centering
\centering

\subfloat[][ \centering Original ] {
\includegraphics[width=0.24\textwidth]{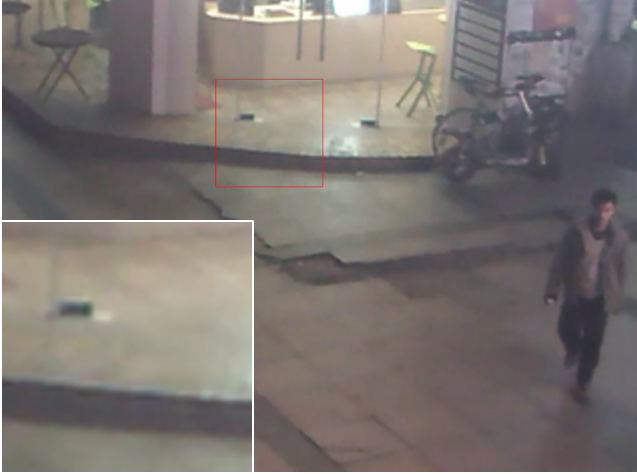}
}
\subfloat[][ \centering Proposed] {
\includegraphics[width=0.24\textwidth]{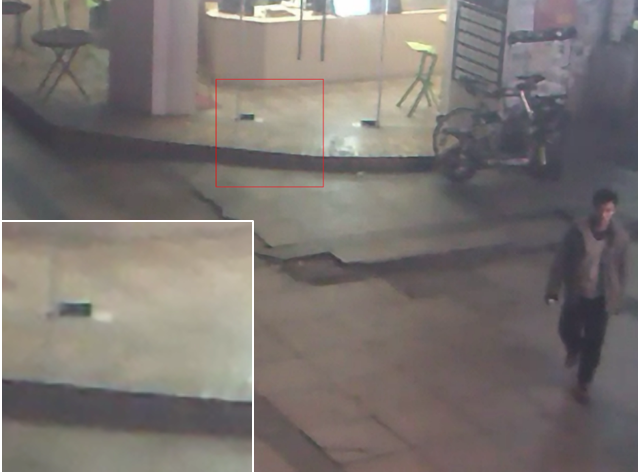}
}
\\

\subfloat[][ \centering Original] {
\includegraphics[width=0.24\textwidth]{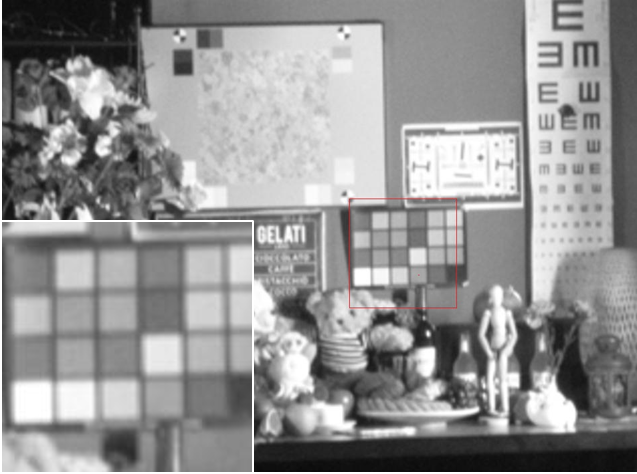}
}
\subfloat[][ \centering Proposed ] {
\includegraphics[width=0.24\textwidth]{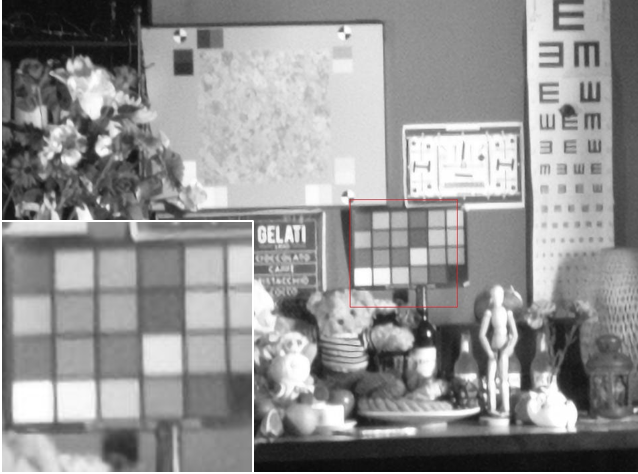}
}
\caption{ Visual results of our model on real-world cases. The upper row shows the case of video surveillance and the lower row shows the case of mobile device. To see clear comparisons, it is better to zoom in the electronic version of this paper. 
}
\label{fig:realworld}
\vspace{-8pt}
\end{figure}

Some promising examples are visualized in Fig.~\ref{fig:set5-visualize} and Fig.~\ref{fig:internet-visualize}. For better viewing, we interpolate the chrominance components by the bicubic method to generate color images. To clearly demonstrate the difference, we choose one 
patch from each image and attach them below. Compared to other methods, our model can produce images with sharper and clearer boundaries. 

\textbf{\textit{Visual Comparison with SRGAN:}} We compare our method with the super-resolution generative adversarial network (SRGAN)~\cite{ledig2016photo}. Because of their proposed adversarial loss, SRGAN has obtained promising performance. However, it still has 
 problems in recovering real details, which is verified by
the comparisons shown in Fig.~\ref{fig:GAN}. It is shown in the enlarged patches of Fig.~\ref{fig:GAN}~(c) and (d) that some waterdrops exist in the ground-truth image disappear, which are produced by SRGAN methods. But these waterdrops are captured by our method and ShCNN. 
As pointed out in~\cite{Mehdi2016enhance}, SRGAN tends to bring in similar textures instead of recovering real details.
Therefore, our proposed framework performs better than SRGAN on recovering more accurate details.

{\textbf{\textit{Discussion on real-world cases:}} To justify the effectiveness of our method, we move one step forward to deal with images from video surveillance and mobile device. Specifically, we apply our model on real-world images with a scaling factor of 3. As reported in Fig.~\ref{fig:realworld}, ``Original'' indicates the original images and ``Proposed'' represent the images processed with our model. As one can observe from results shown in Fig.~\ref{fig:realworld}, ``Proposed'' have fewer artifacts compared with ``Original''. This demonstrates the robustness of our method towards real-world challenges. }

\subsection{Ablation Study}~\label{sec:module_analysis}
In this subsection, we conduct detailed analyses on the proposed modules, i.e., content-adaptive interpolation, BCN and RCN, for better understanding of our framework. 
We hope such analysis can lead to new insights into image restoration researches.

\textit{\textbf{Content-adaptive interpolation}}:
One of the major differences between our model and SRCNN~\cite{dong2014srcnn} is the employment of the deconvolutional layer. To demonstrate the superiority of our design,
we train several fully convolutional networks (FCNs) with various layer numbers for comparisons. Specifically, we increase the number of middle layers from 5 to 16, resulting in FCN-5, FCN-9, FCN-12, and FCN-16. These FCNs follow the bicubic upsampling strategy as in SRCNN~\cite{dong2014srcnn}. Our content-adaptive interpolation consist of 5 convolutional layers and one deconvolutional layer, which contain feature extraction stage, content-adaptive interpolation and BCN. We remove the task of boundary objective to address the effectiveness of content-adaptive interpolation. By comparing content-adaptive interpolation with these FCNs on \textit{Set5} dataset with a scaling factor of 3, we obtain the results shown in Table~\ref{table:LSP-vs-FCN}.

\begin{figure} [ht]\centering
 {
\includegraphics[width=0.41\textwidth]{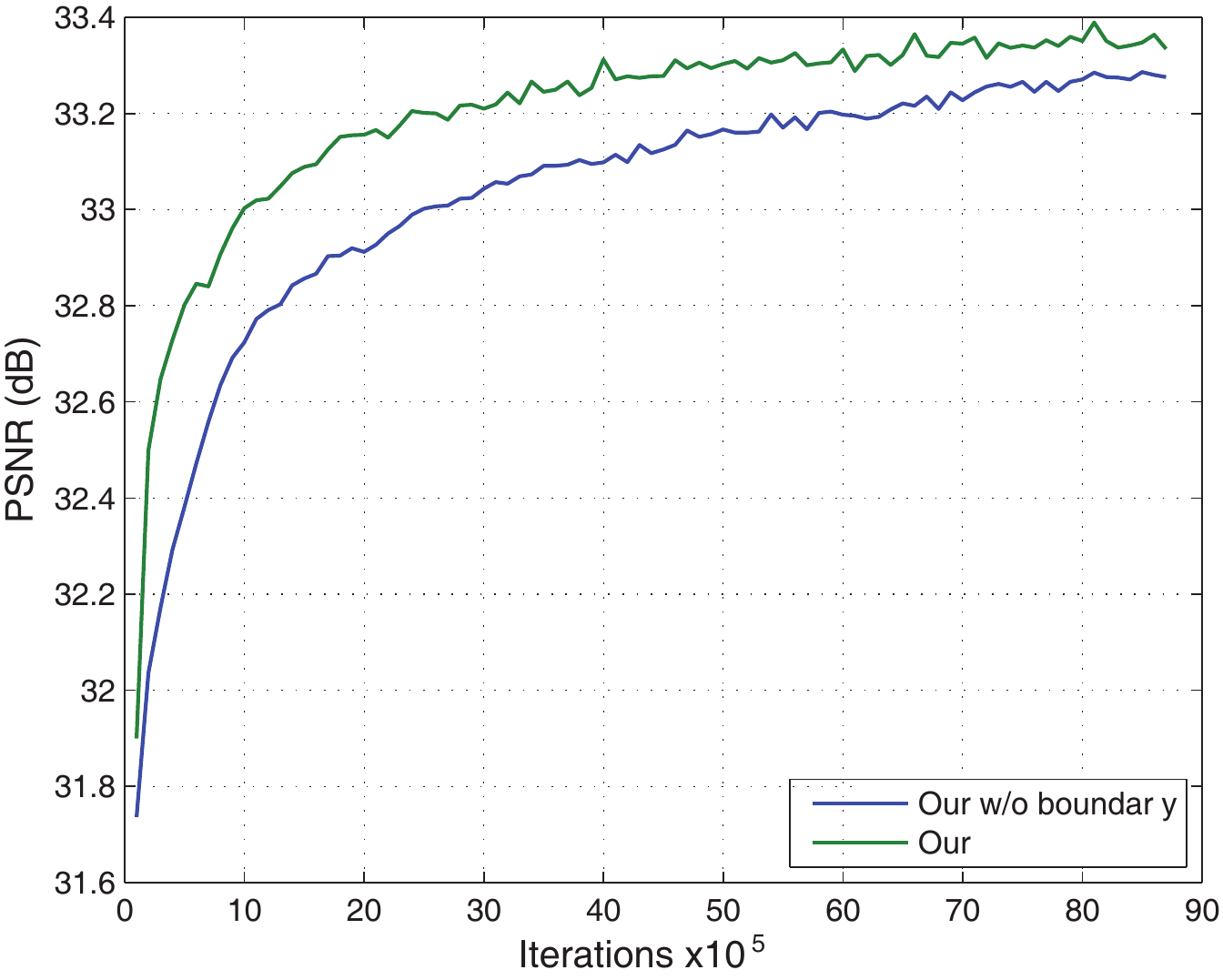} }
\caption{The PSNR curves
generated by models trained with and without edge prediction objective.}
\label{fig:comparison-on-HSP-and-noHSP}
\end{figure}

It is indicated in these results that although the SR performance of FCN keeps increasing as the network depth increases, it still cannot outperform content-adaptive interpolation even when there are 16 layers.
Nevertheless, our content-adaptive interpolation network, which only has 6 layers, surpasses these FCNs by a clear margin. More specifically, content-adaptive interpolation network outperforms FCN-16 by 0.32~dB. This explicitly verifies the superiority of the content-adaptive interpolation.
\begin{table}[t] \centering \footnotesize %
\center
\begin{tabular}{|c|c|c|c|c|c|c|}
\hline
Module  & FCN-5 & FCN-9 & FCN-12 & FCN-16 & LSPM \\ \hline \hline
PSNR (dB) & 32.75 & 32.82 &  32.86  &  32.97  & \bf33.29 \\ \hline
\end{tabular}
\caption{Comparison between content-adaptive interpolation and 
FCNs on \textit{Set5} dataset with a scaling factor of 3. We remove the edge prediction objective to justify the effectiveness of content-adaptive interpolation.}
\label{table:LSP-vs-FCN}
\end{table}

\textit{\textbf{Global Boundary Context}}:
The proposed BCN is motivated by the paradigm of mult-task learning, which incorporates edge estimation as a co-task of HR image generation. Therefore, its analysis is conducted by comparing the SR performance between with and without the edge prediction objective.
Since the \textit{BSD200} dataset contains manually labeled boundary maps, based on which we can easily compute the EPSNR. We compare two profiles of our model on this dataset with a scaling factor of 3 using both PSNR and EPSNR metrics. 
By removing the boundary prediction objective, we degrade BCN into single-task learning and denote it as ``ours w/o boundary''. As illustrated in Table~\ref{table:EPSNR}, the PSNR and EPSNR gains indicate the benefit of multi-task learning. 
Because the boundaries only occupy a small portion of the whole image, the improvement on overall PSNR is minor. However, the large improvement on EPSNR verifies the effectiveness of BCN.
Another benefit of incorporating boundary prediction objective is the acceleration of training process. 
As shown in the PSNR curves of Fig.~\ref{fig:comparison-on-HSP-and-noHSP}, the edge prediction objective not only accelerates the convergence, but also contributes to a higher restoration quality. 
\begin{table}[t] \centering 
\center
\begin{tabular}{|*{3}{c|}}
\hline
Methods & PSNR (dB) & EPSNR (dB) \\
\hline\hline
Bicubic & 27.18 (+0.00) & 22.71 (+0.00) \\
A+~\cite{a+} & 28.36 (+1.21) & 24.28 (+1.57) \\
SRCNN~\cite{dong2014srcnn} & 28.28 (+1.1) & 24.24 (+1.53) \\
SRF~\cite{schulter2015fast} & 28.45 (+1.27) & 24.27 (+1.56) \\
SCN~\cite{wang2015deep} & 28.54 (+1.36) & 24.29 (+1.58) \\
ShCNN~\cite{NIPS2015xuli} & 28.60 (+1.42) & 24.32 (+1.61) \\
\hline  
Ours w/o boundary & 28.68 (+1.46) & 24.36 (+1.65) \\
Ours & \textbf{28.69 (+1.47)} & \textbf{24.43 (+1.72)} \\
\hline
\end{tabular}
\caption{Comparisons on \textit{BSD200} dataset with a scaling factor of 3.}
\label{table:EPSNR}
\end{table}

\textit{\textbf{Local Residue Context}}:
We design RCN to provide complementary information for image SR. Therefore, the SR performance
of our model will be degraded if RCN is removed.
To verify our statement, we use another profile named ``ours w/o RCN'', which is very similar to the previous version of this work~\cite{This_ICME16},  to conduct more comparisons on the aforementioned datasets with a scaling factor of 3. 
Table~\ref{table:dcm} reports the comparison results. It is shown that, although content-adaptive interpolation and BCN can produce HR image of high quality, the SR performance can still be further improved.
The improvement on PSNR is minor because PSNR is a squared error-based metric, which is difficult to reveal subtle structure differences. In contrast, because SSIM concentrates on structure similarity, the improvement on SSIM is more significant.
\begin{table}[t] \centering \small
\center
\begin{tabular}{|*{4}{c|}}
\hline
Test set & Set5 & Set14 & BSD200\\
\hline\hline
Ours w/o RCN & 33.36~dB & 29.57~dB & 28.63~dB \\
Ours & \textbf{33.47}~dB & \textbf{29.64}~dB & \textbf{28.69}~dB\\
\hline
Ours w/o RCN & 0.9162 & 0.8255 & 0.8176 \\
Ours & \textbf{0.9176} & \textbf{0.8273} & \textbf{0.8183} \\
\hline
\end{tabular}
\caption{Comparisons between our model with and without RCN on the PSNR (top) and SSIM (bottom) metrics.}
\label{table:dcm}
\end{table}

\section{Conclusion and Future Work}~\label{sec:con}
In this paper, to address single image super-resolution, we have proposed a novel contextualized multi-task deep learning framework. Our neural network model incorporates global boundary context and residual context to super-resolve images while well preserving their structural details. 
{ Moreover, we have introduced ``content-adaptive interpolation", which leverages a set of filters that are adaptive to the training samples. Different from the kernel estimation in blind image SR which usually employs only a single filter, our proposed content-adaptive interpolation has more filtering parameters and better convenience of being embedded into CNNs.} Our extensive experiments suggest that the proposed method outperforms other leading image super-resolution approaches, and achieves state-of-the-art performances on both popular evaluation metrics and visual quality comparison. 

There are several directions to extend our method. First, we are considering to introduce a perceptual loss into the multi-task optimization, aiming to better capture realistic and meaningful image details. Second, we shall generalize this framework to adapt to video data by taking spatio-temporal coherency into consideration.
Third, considering that additional common knowledge in deep neural networks would be an interesting trial, we intend to utilize complementary spatial-temporal contexts as privileged information for video SR, as suggested by Yang \textit{et al.}~\cite{al2016TMM}.

\section*{Acknowledgements}
This work is partially supported by NSFC (No. 61602533), The Fundamental Research Funds for the Central Universities, in part by Hong Kong Scholars Program and Hong Kong Polytechnic University Mainland University Joint Supervision Scheme. We are grateful to acknowledge NVIDIA for GPU donations.


\bibliographystyle{IEEEbib}
\bibliography{icme2016template}

\begin{IEEEbiography}[{\includegraphics[width=1in,height=1.25in,clip,keepaspectratio]{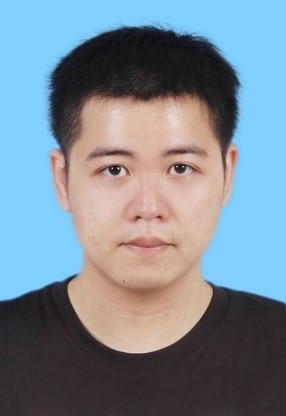}}]{Yukai Shi} received his B.E. degree from Heilongjiang University. He is currently working toward the Ph.D. degree with the School of Data and Computer Science, Sun Yat-Sen University. His research interests including computer vision and machine learning.
\end{IEEEbiography}
\begin{IEEEbiography}[{\includegraphics[width=1in,height=1.25in,clip,keepaspectratio]{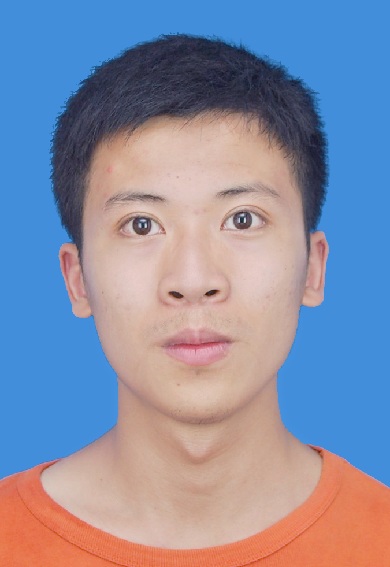}}]{Keze Wang} received his B.S. degree in software engineering from Sun Yat-Sen University, Guangzhou, China, in 2012. He is currently pursuing the dual Ph.D. degree at Sun Yat-Sen University and Hong Kong Polytechnic University, advised by Prof. Liang Lin and Lei Zhang . His current research interests include computer vision and machine learning. More information can be found in his personal website http://kezewang.com
\end{IEEEbiography}

\begin{IEEEbiography}
[{\includegraphics[width=1in,height=1.25in,clip,keepaspectratio]{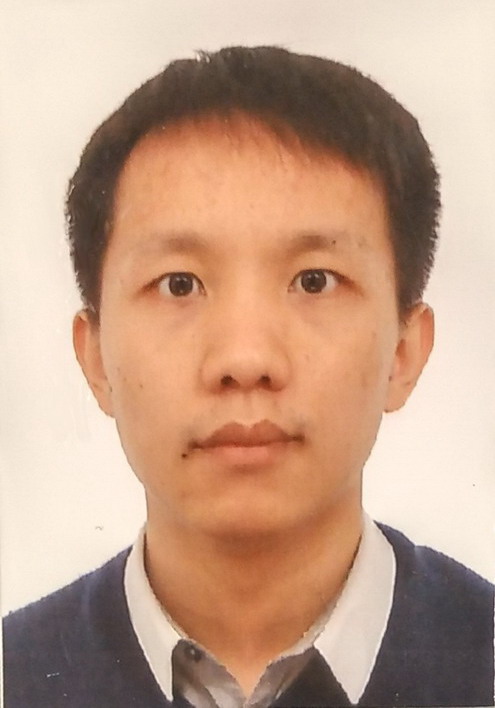}}]{Chongyu Chen} is a distinguished associate research fellow of Sun Yat-sen University. He received his B.S. and Ph.D. degrees from Xidian University, Xi'an, China, in 2008 and 2014, respectively. From 2015 to 2017, he was a post-doctoral fellow at the Hong Kong Polytechnic University. His research interests include image restoration, 3D computer vision, signal separation, etc.
\end{IEEEbiography}

\begin{IEEEbiography}
[{\includegraphics[width=1in,height=1.25in,clip,keepaspectratio]{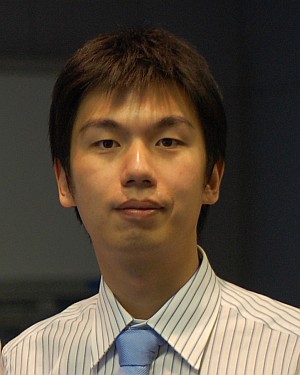}}]{Li Xu} received the BS and MS degrees in computer science and engineering from Shanghai JiaoTong University (SJTU) in 2004 and 2007, respectively, and the PhD degree in computer science and engineering from the Chinese University of Hong Kong (CUHK) in 2010. He joined Lenovo R \& T Hong Kong in Aug 2013, where he leads the imaging \& sensing group in the Image \& Visual Computing (IVC) Lab. Li received the Microsoft Research Asia Fellowship Award in 2008 and the best paper award of NPAR 2012. His major research areas include motion estimation, motion deblurring, image/video analysis and enhancement. He is a member of the IEEE. 
\end{IEEEbiography}

\begin{IEEEbiography}[{\includegraphics[width=1in,height=1.25in,clip,keepaspectratio]{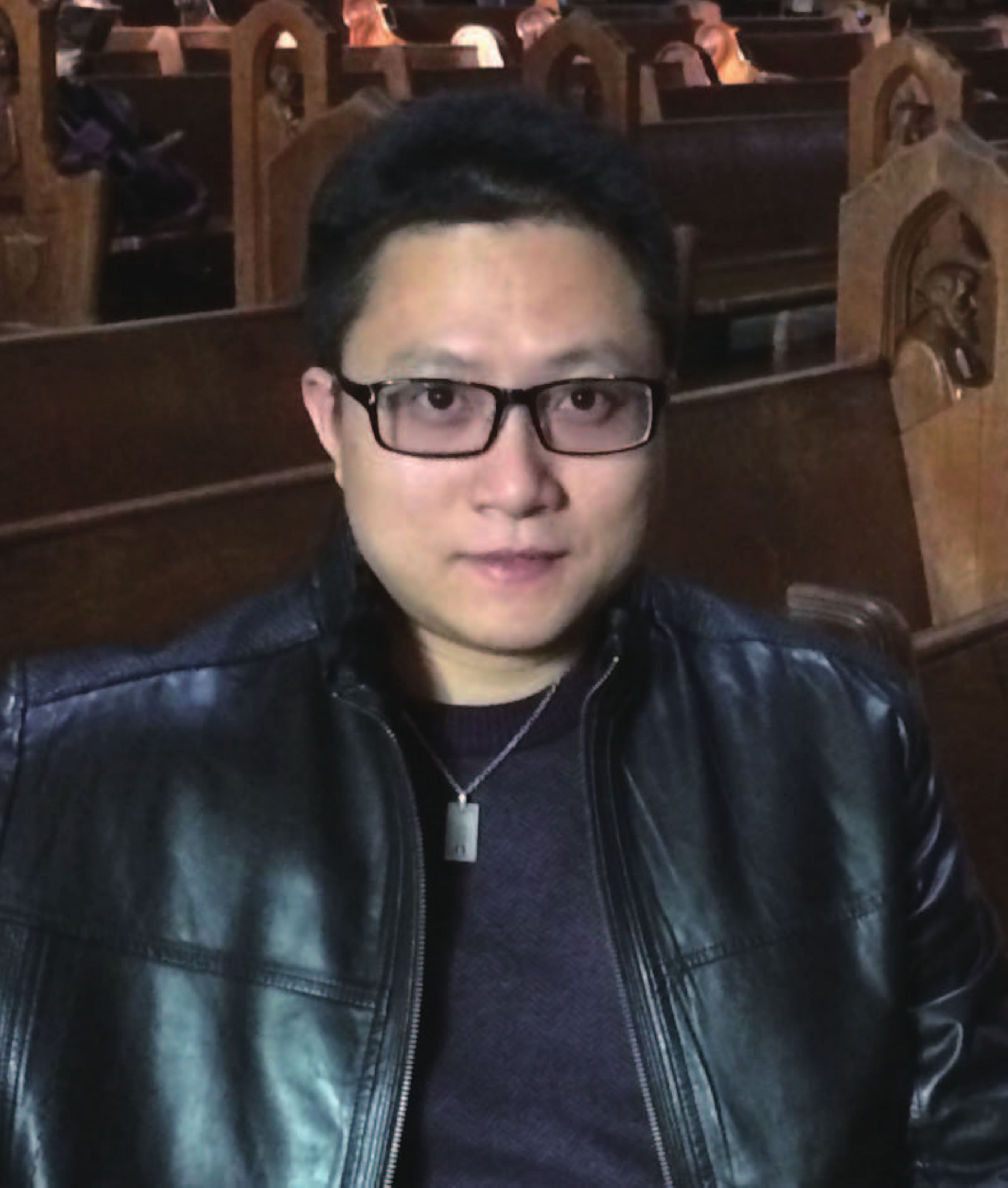}}]{Liang Lin} is a full Professor of Sun Yat-sen University. He received his B.S. and Ph.D. degrees from the Beijing Institute of Technology (BIT), Beijing, China, in 2003 and 2008, respectively, and was a joint Ph.D. student with the Department of Statistics, University of California, Los Angeles (UCLA). From 2008 to 2010, he was a Post-Doctoral Fellow at UCLA. From 2014 to 2015, as a senior visiting scholar he was with The Hong Kong Polytechnic University and The Chinese University of Hong Kong. His research interests include Computer Vision, Data Analysis and Mining, and Intelligent Robotic Systems, etc. He has authorized and co-authorized on more than 100 papers in top-tier academic journals and conferences. He has been serving as an associate editor of IEEE Trans. Human-Machine Systems. He was the recipient of the Best Paper Runners-Up Award in ACM NPAR 2010, Google Faculty Award in 2012, Best Student Paper Award in IEEE ICME 2014, Hong Kong Scholars Award in 2014 and The World's First 10K Best Paper Diamond Award in IEEE ICME 2017. More information can be found in his group website http://hcp.sysu.edu.cn.
\end{IEEEbiography}

\end{document}